\documentclass[runningheads,a4paper]{llncs}
\usepackage{graphicx}
\usepackage{amsmath}
\usepackage{amssymb}
\usepackage[utf8]{inputenc}
\usepackage{url}
\usepackage{algorithm2e}
\begin{document}
\frontmatter          
\pagestyle{headings}  

\title{On estimating total time to solve SAT in distributed computing environments:}
\subtitle{Application to the SAT@home project}
\titlerunning{On estimating total time to solve SAT}  
\author{Alexander Semenov \and Oleg Zaikin}
\authorrunning{A.\,Semenov \and O.\,Zaikin}
\institute{Institute for System Dynamics and Control Theory SB RAS, Irkutsk, Russia
\email{biclop.rambler@yandex.ru, zaikin.icc@gmail.com}}

\maketitle              

\begin{abstract}
This paper proposes a method to estimate the total time required to solve SAT in distributed environments via partitioning approach. It is based on the observation that for some simple forms of problem partitioning one can use the Monte Carlo approach to estimate the time required to solve an original problem. The method proposed is based on an algorithm for searching for partitioning with an optimal solving time estimation. We applied this method to estimate the time required to perform logical cryptanalysis of the widely known stream ciphers A5/1 and Bivium. The paper also describes a volunteer computing project SAT@home aimed at solving hard combinatorial problems reduced to SAT. In this project during several months there were solved 10 problems of logical cryptanalysis of the A5/1 cipher that could not be solved using known rainbow tables.
\end{abstract}

\keywords{volunteer computing, BOINC, partitioning, Monte Carlo method, predictive function, tabu search, A5/1, Bivium, SAT@home}
\section{Introduction}
In recent years, solving large scale computational problems via volunteer computing projects gained a lot of popularity. Nowadays there are about 70 active projects, the majority of them are based on the BOINC platform \cite{DBLP:conf/grid/Anderson04}. Total performance of all BOINC projects is more than 7 petaflops. The most important results obtained in volunteer computing projects include the discovery of new pulsars in the Einstein@home, and of large prime numbers of a special kind in the PrimeGrid project. 

Volunteer computing is a type of distributed computing. Actually a volunteer computing project is a desktop grid constructed from PCs of private persons called volunteers. It is important to note that volunteers contribute resources of their computers for free, so they assume no obligation to the organizers of the project. Therefore, a list of active project participants (and consequently a structure of a desktop grid) can vary greatly during the computational experiment. PCs of volunteers receive tasks from server, process them and send results back to the server. In volunteer projects tasks should be solved independently because volunteer PCs can communicate only with project server, moreover do it rarely and irregularly. In general, for a project to work effectively it should contain the following components: stable 24/7 server, internet site with the goal of the project clearly pointed out and a number of client applications for various computing platforms. 

It is well-known that a lot of important combinatorial problems (for example from areas of formal verification, planning or bioinformatics) can be effectively reduced to SAT \cite{DBLP:series/faia/2009-185}. Despite quite significant progress in the development of SAT solvers there remain hard SAT instances that cannot be solved without the use of large amounts of computational resources of various types. Therefore, in our opinion, it is essential to use volunteer computing for solving hard SAT instances. We develop and maintain a volunteer computing project SAT@home \cite{website:sat@home} specially designed for solving SAT problems via partitioning approach. 

In volunteer computing projects, excluding the projects with ambitious goals like SETI@home, it is very important to know how much time it will take to solve a particular problem. It is considered to be normal if a time estimation involves months or even years. Knowledge about this time provides volunteers with an additional motivation by showing how effectively they progress in solving the problem. 

Further we present a Monte Carlo method of estimating time to solve SAT problems in distributed computing environments. For a given solver and a fixed partitioning of the original SAT problem this method statistically estimates values of several qualitative parameters of the chosen partitioning. One of these parameters corresponds to the time required to solve the considered SAT problem. To automatically search for a partitioning with minimal estimation of time we use a tabu search based algorithm. Our method was used to estimate the time required for solving of several hard SAT instances in the SAT@home project.

A brief outline of the paper is given below. In the next section we present a Monte Carlo method that makes it possible to estimate time required for solving of SAT problems. Section 3 contains some implementation details and the results of computational experiments that show practical applicability of the proposed method. In Section 4 we briefly describe the volunteer computing project SAT@home and present some of the results obtained in this project. In Section 5 we consider related works. 

\section{Monte Carlo method for estimating time of solving of SAT problem via partitioning approach}

In \cite{Hyvarinen11,DBLP:conf/sat/HyvarinenJN06,DBLP:conf/lpar/HyvarinenJN10,DBLP:conf/cp/HyvarinenJN11} various approaches to partitioning SAT problems were studied. Further we will use the notation of \cite{Hyvarinen11}. Consider an arbitrary SAT-problem for CNF $C$ over a set of Boolean variables $X$. Partitioning of $C$ is a set of formulae 
$$
C \cdot F_i, i \in \left\{1, \ldots,S \right\},
$$
such that for any $i,j:i \ne j$, formula $C\cdot F_i \cdot F_j$ is unsatisfiable and 
$$
C\equiv C\cdot F_1\vee\ldots\vee C\cdot F_S.
$$

When one has a partitioning of the original SAT problem, satisfiability problems for $C\cdot F_j$, $j \in\left\{1,\ldots,S\right\}$, can be solved independently in a distributed computing environment. There exist various partitioning techniques. For example, one can construct $\left\{F_j\right\}_{j=1}^S$ using a scattering procedure \cite{DBLP:conf/sat/HyvarinenJN06}, a guiding path solver \cite{DBLP:journals/jsc/ZhangBH96} or a lookahead solver \cite{DBLP:conf/hvc/HeuleKWB11}, \cite{DBLP:conf/lpar/HyvarinenJN10}.
Unfortunately, in general case for these partitioning methods it is not possible to estimate the total time required for solving the original SAT problem. 

However, a partitioning method that makes it possible to construct such estimations was used in a number of papers about solving cryptanalysis problems via SAT approach \cite{DBLP:conf/sat/EibachPV08}, \cite{Mcdonald_attackingbivium}, \cite{Ecrypt_TOOLS10b}, \cite{DBLP:conf/sat/SoosNC09}.
According to this method, from the set of variables $X$ of an original CNF $C$ we choose a subset $\tilde{X}$ and consider a family consisting of all the CNFs that are produced from $C$ by substituting all the $2^{|\tilde{X}|}$ truth assignments of variables from $\tilde{X}$. It is clear, that in this case the formula $F_{1} \vee F_{2}, \vee \ldots, \vee F_{S}$ is a DNF in which $S=2^{|\tilde{X}|}$, and at the first glance it may look like a significant drawback of the approach. On the other hand one can solve $N$, $N<<2^{|\tilde{X}|}$ SAT problems for CNFs randomly chosen from the family considered, calculate the average time of their solving and, based on this information, estimate the total time required for solving the original SAT problem. Below we show that this method can be formally justified using the Monte Carlo approach. We also propose an algorithm that searches for a partitioning with a minimal estimated time and show practical applicability of this procedure to the inversion problems of several cryptographic functions.

\subsection{Monte Carlo approach to statistical estimation of quality of partitioning of SAT problem}

Consider a SAT problem for an arbitrary CNF formula $C=C(X)$ over the set of Boolean variables $X=\left\{x_1,\ldots,x_n\right\}$. We refer to an arbitrary $\tilde{X}=\left\{ x_{i_1},\ldots,x_{i_d}\right\}$, $\tilde{X}\subseteq X$, $\left\{i_1,\ldots,i_d\right\}\subseteq \left\{1,\ldots,n\right\}$, as a decomposition set for the SAT problem considered. Further a CNF formula obtained as a result of substituting a truth assignment $x_{i_1}=\alpha_1,\ldots,x_{i_d}=\alpha_d$ to $C$ is denoted as $C\left[\tilde{X}/(\alpha_1,\ldots,\alpha_d)\right]$. A set of CNFs
$$
\Delta\left( C,\tilde{X} \right)= \left\{ C\left[\tilde{X}/(\alpha_1,\ldots,\alpha_d)\right] \right\} _{(\alpha_1,\ldots,\alpha_d)\in\{0,1\}^d} 
$$
is called a decomposition family produced by $\tilde{X}$. It is easy to see that in accordance with the above $\Delta\left(C,\tilde{X}\right)$ is a partitioning of $C$.

Consider some algorithm $A$ solving SAT. In the remainder of the paper we presume that $A$ is complete, i.e. its runtime is finite for an arbitrary input. We also presume that $A$ is a deterministic algorithm that does not involve randomization. We need the latter condition to use the Monte Carlo approach correctly. 

We denote an amount of time required by $A$ to solve all the CNFs from $\Delta\left(C,\tilde{X}\right)$ as $t_A\left(C,\tilde{X}\right)$. Below we mainly concentrate on estimation problem for $t_A\left(C,\tilde{X}\right)$.

Let's define a uniform distribution on the set $\{0,1\}^d$. With each randomly chosen vector $\left(\alpha_1,\ldots,\alpha_d\right)$ from $\{0,1\}^d$ we associate a value
$$
\xi_A\left(\alpha_1,\ldots,\alpha_d\right),
$$
that is equal to the time required by algorithm $A$ to solve SAT for formula $ C\left[\tilde{X}/(\alpha_1,\ldots,\alpha_d)\right]$. Therefore, a random variable
$$
\xi_A\left( C,\tilde{X}\right)=\left\{ \xi_A \left( \alpha_1,\ldots,\alpha_d\right)\right\}_{(\alpha_1,\ldots,\alpha_d)\in\{0,1\}^d}
$$
with some probability distribution is defined. Due to the completeness of $A$, variable $\xi_A\left(C,\tilde{X}\right)$ has a finite expected value $\mathrm{E}\left[\xi_A\left(C,\tilde{X}\right) \right]$ and a finite variance $\mathrm{Var}\left(\xi_A\left(C,\tilde{X}\right) \right)$. It is important, that since we presume that $A$ is a deterministic algorithm, then $N$ independent observations of values of $\xi_A\left(C,\tilde{X}\right)$ can be considered as a single observation of $N$ independent random variables with the same distribution as $\xi_A\left(C,\tilde{X}\right)$.

It is not difficult to prove that
\begin{equation}
\label{eq1}
t_A\left(C,\tilde{X}\right)=2^d\cdot \mathrm{E} \left[ \xi_A\left(C,\tilde{X}\right) \right].
\end{equation} 
Below we refer to $\tilde{X}\in 2^X$ with a minimal value $t_A\left(C,\tilde{X}\right)$ as an optimal decomposition set.
	
To estimate the value of $\mathrm{E}\left[\xi_A\left(C,\tilde{X}\right) \right]$ we will use the Monte Carlo method \cite{Kalos:109491}, \cite{Metropolis49}. According to this method, in order to approximately calculate the expected value $\mathrm{E}[\xi]$ of an arbitrary random variable $\xi$ a probabilistic experiment is used, that consists of $N$ independent observations of values of $\xi$. Let $\xi^1,\ldots, \xi^N$ be results of the corresponding observations. They can be considered as a single observation of $N$ independent random variables with the same distribution, i.e. the following equalities hold:
$$
\mathrm{E}[\xi]=\mathrm{E}[\xi^1]=\ldots=\mathrm{E}[\xi^N], \mathrm{Var}(\xi)=\mathrm{Var}(\xi^1)=\ldots=\mathrm{Var}(\xi^N).
$$
If $\mathrm{E}[\xi]$ and $\mathrm{Var}(\xi)$ are both finite then from Central Limit Theorem \cite{feller-vol-2} we have the main formula of the Monte Carlo method
\begin{equation}
\label{eq2}
\mathrm{Pr}\left\{ \left| \frac{1}{N}\cdot \sum\limits_{j=1}^{N}\xi^j-\mathrm{E}[\xi]\right|<\frac{\delta_\gamma \cdot \sigma}{\sqrt{N}} \right\}=\gamma.
\end{equation}
Here $\sigma=\sqrt{\mathrm{Var}(\xi)}$ stands for a standard deviation, $\gamma$ --- for a confidence level, $\gamma=\mathrm{\Phi}(\delta_\gamma)$, where $\mathrm{\Phi}(\cdot)$ is the normal cumulative distribution function. It means that under the considered assumptions the value 
\[
\bar{\xi}=\frac{1}{N}\cdot\sum\limits_{j=1}^N\xi^j
\]
 is a good approximation of $\mathrm{E}[\xi]$, when the number of observations $N$ is large enough. For any given $N$ the quality of this approximation depends on the value of $\mathrm{Var}(\xi)$. In practice to estimate $\mathrm{Var}(\xi)$ an unbiased sample variance 
\begin{equation}
\label{eq3}
s^2=\frac{1}{N-1}\cdot \sum\limits_{j=1}^N\left(\xi^j-\bar{\xi}\right)^2
\end{equation}				
is used. In this case instead of \eqref{eq2} a following formula is applied \cite{Wilks62}
\begin{equation}
\label{eq4}
\mathrm{Pr}\left\{ \left| \frac{1}{N}\cdot \sum\limits_{j=1}^{N}\xi^j-\mathrm{E}[\xi]\right|<\frac{t_{\gamma,N-1}\cdot s}{\sqrt{N}} \right\}=\gamma,
\end{equation}
where $t_{\gamma,N-1}$ is a quantile of a Student's distribution with $N-1$ degrees of freedom that corresponds to the confidence level $\gamma$ . If for example $\gamma=0,999$ and $N\geq 10000$ then $t_{\gamma,N-1}\approx 3.29$.

In our case it is important to note that $N$ can be significantly less than $2^d$. It means that the preprocessing stage can be used to estimate the total time, required for processing the whole decomposition family $\Delta\left(C,\tilde{X}\right)$.

So the process of approximate calculating of value \eqref{eq1} for a given $\tilde{X}$ is as follows. We randomly choose $N$ truth assignments of variables from $\tilde{X}$ and denote this set as: 
\begin{equation}
\label{eq5}
\Theta \left(\tilde{X} \right) = \left\{ \left( \alpha_1^1,\ldots,\alpha_d^1\right),\ldots,\left(\alpha_1^N,\ldots,\alpha_d^N \right) \right\}.
\end{equation}
Consider random variables 
$$
\xi^j=\xi_A\left(\alpha_1^j,\ldots,\alpha_d^j\right), j=1,\ldots,N,
$$
and calculate the value
$$
F_{A,C}\left(\tilde{X}\right)=2^d\cdot\left( \frac{1}{N}\cdot\sum\limits_{j=1}^N \xi^j \right).
$$
By the above if $N$ is large enough then the value $F_{A,C}\left(\tilde{X}\right)$ can be considered as a good approximation of \eqref{eq1}.

Below we refer to function $F_{A,C}\left(\cdot\right)$ as a predictive function. Note that values of the predictive function can be calculated using a usual PC or a computing cluster, given a qualitative random number generator. Therefore, instead of searching for an optimal decomposition set one can search for decomposition set with minimal value of predictive function.

\subsection{Algorithm for predictive function minimization}
Despite quite a natural formulation, the problem described has some specific features.

\begin{enumerate}
	\item The value of $F_{A,C}\left(\tilde{X}\right)$ cannot be effectively calculated for an arbitrary $\tilde{X}$ since it is easy to construct such small $\tilde{X}$ that the time required for calculating $F_{A,C}\left(\tilde{X}\right)$ is comparable with the time required for solving the original SAT problem.
	
	\item The value of $F_{A,C}\left(\tilde{X}\right)$ represents the reaction of computing environment to the corresponding decomposition set. Therefore, methods relying on the analytical properties of an objective function cannot be applied to the problem of minimizing $F_{A,C}\left(\cdot\right)$.
	
\end{enumerate}

Because of these features the most natural minimization strategy for $F_{A,C}\left(\cdot\right)$ is a strategy of ``successive improvements''. It implies that at the first step we should construct an initial decomposition set $\tilde{X}_0$ for which the value $F_{A,C}\left(\tilde{X}_0\right)$ can be calculated in a short time. After this, we try to improve this value by observing the neighbourhood of a point corresponding to $\tilde{X_0}$ in some search space $\Re$ . Thus, the process of minimization consists in moving from point to point looking for the one with minimal $F_{A,C}\left(\cdot\right)$. 

In some minimization algorithms it is allowed to calculate a value of an objective function in the same point of a search space more than once. It is feasible for problems where the objective function can be calculated easily in any point of $\Re$. Due to the reasons mentioned above, in our case it is undesirable. Ideally the calculation of $F_{A,C}\left(\cdot\right)$ in an arbitrary point of $\Re$ shouldn't be performed more than once. In the algorithm described below we keep all points for which the value of $F_{A,C}\left(\cdot\right)$ is already calculated. It naturally corresponds to the basic idea of tabu search (TS) \cite{Glover:1997:TS:549765}.

Let's start from defining the search space. Note that an arbitrary set $\tilde{X}\in 2^X$ can be described using a Boolean vector 
\begin{equation}
\label{eq6}
\chi\left(\tilde{X}\right)=\chi=\left(\chi_1,\ldots,\chi_n\right),\chi_i=\left\{\begin{array}{l}
1, x_i\in\tilde{X}\\
0, x_i\notin\tilde{X}
\end{array} \right., i=1,\ldots,n.
\end{equation}
Then it is convenient to define the search space $\Re$ as an $n$-dimensional Boolean hypercube $E^n=\left\{0,1\right\}^n$. For an arbitrary point $\chi\in E^n$ a neighbourhood $\mathcal{N}_\rho\left(\chi\right)$ of radius $\rho$ is defined as a set of vectors $\chi'$ from $E^n$ such that
$$
dist_H\left(\chi',\chi\right)\leq \rho,
$$
where $dist_H\left(\chi_1,\chi_2\right)$ stands for the Hamming distance between $\chi_1$ and $\chi_2$. Punctured neighbourhood of point $\chi$ is a set 
$$
\mathcal{N}_\rho^*=\mathcal{N}_\rho\left(\chi\right)\backslash \{\chi\}.
$$
Further by $F\left(\chi\right)$ we denote $F_{A,C}\left(\tilde{X}\right)$, where $\chi=\chi\left(\tilde{X}\right)$ is in accordance with \eqref{eq6}. We consider the problem of search for a minimum of the function $F\left(\cdot\right)$ over $E^n$. 

A simple local search \cite{DBLP:books/ph/PapadimitriouS82} stops after finding a local extremum. There are various techniques that make it possible to escape from such points. In accordance with the main TS principle we should move from the local extremum to some point that is not included in a current tabu list $T$ . It may occur that the value of the objective function in a new point is worse than its best known value. After we move to a new point a local search stage is launched in a punctured neighbourhood of this point. 

Types of constraints in the tabu list and ways of their usage differ from problem to problem and are determined based on a problem's individual features. In our approach it is convenient to keep all the points we have checked in the form of two lists of constraints: $L_1$ and $L_2$. List $L_1$ stores points $\chi\in E^n$ such that for all $\chi{'}\in \mathcal{N}_\rho\left(\chi\right)$ we have already calculated $F\left(\chi{'}\right)$. In list $L_2$ we keep points $\chi\in E^n$ such that $F\left(\chi\right)$ have been calculated and there exists $\chi{'}\in \mathcal{N}_\rho\left(\chi\right)$ for which we haven't calculated $F\left(\chi{'}\right)$ yet. To reflect this information every point in $L_2$ is represented by two vectors: a vector $\chi\in E^n$ and a Boolean vector $\theta\left(\chi\right)$ of length $\sum\limits_{i=1}^\rho
{\binom{n}{i}}$. Vector $\theta\left(\chi\right)$ stores information about points from $\mathcal{N}_\rho\left(\chi\right)$: components of $\theta\left(\chi\right)$ equal to $1$ correspond to the points in which the calculation of $F\left(\cdot\right)$ has already been performed (other components are equal to $0$). Vector $\theta\left(\chi\right)$ is referred to as a neighbourhood vector of $\chi$.

Below we give a brief outline of our algorithm. The algorithm is split into iterations. Denote iteration with number $t\geq 0$ as $I\left(t\right)$ and the best known value of $F\left(\cdot\right)$ obtained at this iteration as $\Psi_t$ . Also, at the start of every iteration we have a current point $\chi_t^c$ that is obtained as a result of the previous iteration. The algorithm starts from some point $\chi_0$ in which the value of predictive function can be calculated fast. We will discuss possible ways of choosing $\chi_0$ later. At the start of $I\left(0\right)$ we assume that $\chi_0^c=\chi_0$, $\Psi_0=F\left(\chi_0\right)$, $L_2=\left\{\chi_0\right\}$, $L_1=\emptyset$. Further we describe the transition from $I\left(t\right)$, $t\geq 0$, to $I\left(t+1\right)$. 

At the beginning of $I\left(t\right)$ we know point $\chi_t^c$ and its neighbourhood vector $\theta\left(\chi_t^c\right)$. We start the local search stage at $\mathcal{N}_\rho^*\left(\chi_t^c\right)$. In particular we consequently check points $\chi{'}\in \mathcal{N}_\rho^*\left(\chi_t^c\right)$ that correspond to zero components of $\theta\left(\chi_t^c\right)$. We calculate value $F\left(\chi{'}\right)$ and add $\chi{'}$ to $L_2$. Every time when we add a new point $\chi{'}$ to $L_2$ we have to modify the constraints in this list: for all $\chi{''}\in L_2$ such that $dist_H\left(\chi{'},\chi{''}\right)\leq \rho$ we set a component of $\theta\left(\chi{''}\right)$, corresponding to $\chi{'}$, to be $1$. Neighbourhood vector $\theta\left(\chi{'}\right)$ is modified accordingly. If for some point from $L_2$ its neighbourhood vector consists only of 1s then this point is removed from $L_2$ and added to $L_1$. If for all $\chi{'}\in \mathcal{N}_\rho^*\left(\chi_t^c\right):F\left(\chi{'}\right)\geq \Psi_{t-1}$ then $\Psi_t=\Psi_{t-1}$, and as $\chi_{t+1}^c$ we choose a point from $L_2$ according to some heuristic, iteration $I\left(t\right)$ ends and iteration $I\left(t+1\right)$ starts. If $F\left(\chi{'}\right)<\Psi_{t-1}$ and $\theta\left(\chi{'}\right)$ contains at least one 0 then $\Psi_t=F\left(\chi{'}\right)$, $\chi_{t+1}^c=\chi{'}$, iteration $I\left(t\right)$ ends and iteration $I\left(t+1\right)$ starts. If $F\left(\chi{'}\right)<\Psi_{t-1}$ but $\theta\left(\chi{'}\right)$ contains only 1s then $\Psi_t=F\left(\chi{'}\right)$, as $\chi_{t+1}^c$ we choose a point from $L_2$ according to some heuristic, iteration $I\left(t\right)$ ends and iteration $I\left(t+1\right)$ starts.

The algorithm stops if either list $L_2$ becomes empty, i.e. all the points from this list are moved to $L_1$, or the time limit is exceeded. It is easy to see that during the work of the algorithm a calculation of the value of the predictive function in an arbitrary point of $E^n$ is performed at most once. A decomposition set that corresponds to the best known value of the predictive function at the moment when algorithm stops we denote as $\tilde{X}_*$.

In computational experiments (see Section 3) we used $\rho=1$ and a following heuristic to choose a current point from $L_2$: it is randomly chosen from the points with the minimal Hamming distance to the point with the best known value of the predictive function.

\subsection{Additional improvements of the predictive function minimization algorithm}

Here we present a technique that makes it possible to significantly speed up the algorithm proposed. 

It is easy to see that the majority of time required to compute $F_{A,C}\left(\tilde{X}\right)$ is spent on calculation of values $\xi^j=\xi_A\left(\alpha_1^j,\ldots,\alpha_d^j\right),j=1,\ldots,N$. However, the calculation of $F_{A,C}\left(\tilde{X}\right)$ can be organized as a following iterative process: 
\begin{equation}
\label{eq7}
F_{A,C}^1\left(\tilde{X}\right)=\frac{2^d}{N}\cdot\xi^1, F_{A,C}^j\left(\tilde{X}\right)=F_{A,C}^{j-1}\left(\tilde{X}\right)+\frac{2^d}{N}\cdot\xi^j,j=2,\ldots,N.
\end{equation}
Thus, 
$$
F_{A,C}\left(\tilde{X}\right)=F_{A,C}^N\left(\tilde{X}\right).
$$
It should be noted that the order of summation in \eqref{eq7} is insignificant. Assume that we need to calculate the value of $F_{A,C}\left(\cdot\right)$ in some point $\chi\left(\tilde{X}'\right)$ and $\Psi$ is our current best known value. Suppose that for some $k<N$ values $\xi^{i_1},\ldots,\xi^{i_k}$, $\left\{i_1,\ldots,i_k\right\} \subset \left\{1,\ldots,N\right\}$, have already been obtained and inequality 
\[
\frac{2^d}{N}\cdot \sum\limits_{r=1}^k \xi^{i_r}>\Psi
\]
holds. Then it is clear that $F_{A,C}\left(\tilde{X}'\right)>\Psi$. In this situation we can interrupt the process of calculation of $F_{A,C}\left(\tilde{X}'\right)$ and move to the next point of the search space. We note that in this case the correctness of the algorithm described is not affected.

An ability of the algorithm proposed to construct a good decomposition set in relatively short amount of time greatly depends on the choice of $\tilde{X}_0$. As we already mentioned earlier we should choose $\tilde{X}_0$ in such a way that $F_{A,C}\left(\tilde{X}_0\right)$ is calculated fast. In a general case we can always assume $\tilde{X}_0=X$ . However, for many SAT problems it is possible to choose $\tilde{X}_0\subset X$ such that $\left|\tilde{X}_0\right|<<\left|X\right|$ and, nevertheless, the value $F_{A,C}\left(\tilde{X}_0\right)$ can be computed effectively. In particular for a SAT problem that encodes the inversion problem of a cryptographic function we can choose $\tilde{X}_0$ as a corresponding Strong Unit Propagation Backdoor Set (SUPBS, \cite{DBLP:journals/constraints/JarvisaloJ09}). In this case we also can search for $\tilde{X}_*$ only among the subsets of $\tilde{X}_0$ , $\tilde{X}_0\subset X$, decreasing the power of the search space to $2^{|\tilde{X}_0|}$.

\section{Implementation and computational experiments}

The algorithm described in Section 2 was implemented as a parallel program \textsc{pdsat} that uses the MPI library. One MPI process of \textsc{pdsat} is assigned to be a master process, others --- to be slave processes. For each new point $\tilde{X}=\left\{x_{i_1},\ldots,x_{i_d}\right\}$ of the search space the master process constructs a set of vectors $\Theta\left(\tilde{X}\right)\subseteq \{0,1\}^d$, $\left|\Theta\left(\tilde{X}\right)\right|=N$ (see \eqref{eq5}), using Mersenne twister pseudorandom number generator. After receiving $\left(\alpha_{i_1}^j,\ldots,\alpha_{i_d}^j\right)\in\Theta\left(\tilde{X}\right)$, $j\in\{1,\ldots,N\}$, a slave process starts solving the SAT problem for CNF $C\left[\tilde{X}/\left(\alpha_{i_1}^j,\ldots,\alpha_{i_d}^j\right)\right]$. Below the set $\left\{ C\left[\tilde{X}/\left(\alpha_{i_1}^j,\ldots,\alpha_{i_d}^j\right)\right]\right\}_{j=1}^N$ is called a sample for a decomposition set $\tilde{X}$.

The interruption technique described in Subsection 2.3 was implemented in \textsc{pdsat}. The master process tracks the total time spent on processing the set $\Theta\left(\tilde{X}\right)$ by all the slave processes. If it decides that, according to \eqref{eq7}, the value of the predictive function for $\tilde{X}$ will exceed its best known value then the master process interrupts the processing $\Theta\left(\tilde{X}\right)$ by sending asynchronous messages to the slave processes. 

A sequential SAT-solver underlies every slave process. In our experiments we used \textsc{MiniSat-C 1.14.1} and \textsc{MiniSat 2.2} \cite{DBLP:conf/sat/EenS03}.

Let's present the results of computational experiments on constructing decomposition sets related to the SAT problems that encode the inversion of widely known cryptographic functions --- A5/1 and Bivium. In two experiments described further \textsc{pdsat} was taking 80 cores of a computing cluster. In both cases \textsc{pdsat} stopped because of reaching the timeout of 4 days. For each new point of the search space \textsc{pdsat} processed a sample of 10 000 CNFs ($N=10000$ ).

The A5/1 keystream generator consists of 3 LFSRs (linear feedback shift register \cite{DBLP:books/crc/MenezesOV96}) that are shifted asynchronously. This generator was described in details in \cite{DBLP:conf/fse/BiryukovSW00}. Cryptanalysis of the A5/1 generator consists in finding the initial contents of LFSRs (64 bits) based on the known keystream fragment.

\begin{figure}[ht]
\centering
\includegraphics[height=5cm]{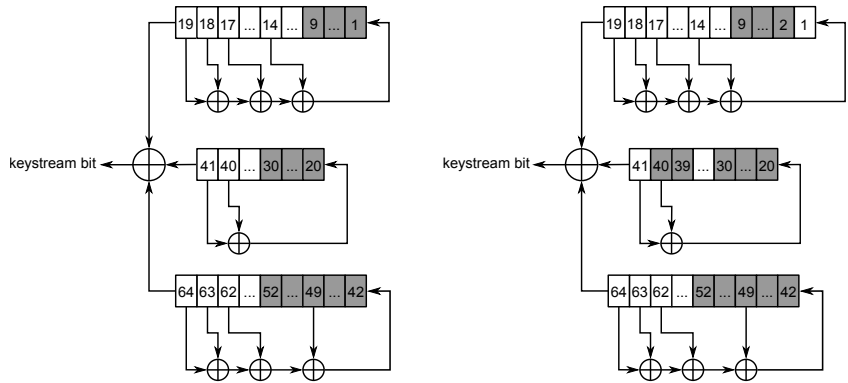}
\caption{Decomposition sets for logical cryptanalysis of A5/1: the one from  \cite{DBLP:conf/pact/SemenovZBP11} (left-hand side) and the one found by \textsc{pdsat} (right-hand side)}
\label{a5-1_sets}
\end{figure}

Usually if the cryptanalysis is considered as a SAT problem then it is called a logical cryptanalysis \cite{DBLP:journals/jar/MassacciM00}. Logical cryptanalysis of the A5/1 generator with first 144 bits of the keystream known was described in \cite{DBLP:conf/pact/SemenovZBP11} where a decomposition set of 31 variables was found manually, guided by the features of the A5/1 algorithm. LFSRs cells corresponding to the variables from this set are marked with grey in the scheme of the A5/1 generator on the left-hand side of Fig. \ref{a5-1_sets}. For further convenience we enumerate cells of the A5/1 registers using continuous numbering (and do the same for Bivium later).

On the right-hand side of Fig. \ref{a5-1_sets} we present a decomposition set for logical cryptanalysis of A5/1 that was found automatically by \textsc{pdsat}. As an $\tilde{X}_0$ \textsc{pdsat} was given a SUPBS of a CNF encoding the A5/1 cryptanalysis problem (64 variables corresponding to the initial state of A5/1 registers). $\tilde{X}_*$ was constructed as a subset of $\tilde{X}_0$.

Best results for the problem of logical cryptanalysis of the A5/1 generator were obtained using \textsc{MiniSat-C 1.14.1} solver with light modifications. Description of these modifications was presented in \cite{DBLP:conf/pact/SemenovZBP11}. Table \ref{a5-1_results} presents some qualitative parameters of decomposition sets from Fig. \ref{a5-1_sets}. Min. time, Max. time and Avg. time stand for the minimal, maximal and average time (in seconds) taken to solve SAT problems for CNFs from the sample, $s^2$ stands for the unbiased sample variance (see \eqref{eq3}), $F\left( \cdot \right)$ --- for the value of predictive function. Values in Table \ref{a5-1_results} are calculated for one core of Intel Xeon E5450 processor.

\begin{table}
\caption{Qualitative parameters of decomposition sets from Fig. \ref{a5-1_sets}. }
\label{a5-1_results}
\centering
\begin{tabular}{p{2.2 cm}|p{1.75cm}|p{1.75cm}|p{1.75cm}|p{1.75cm}|p{1.75cm}}

Sets& Min. time & Max. time & Avg. time & $s^2$ & $F\left( \cdot \right)$\\
\hline 
\hline
from \cite{DBLP:conf/pact/SemenovZBP11} & 0.00020 & 2.38342 & 0.21020 & 0.02359 & 4.45140e+08 \\
(31 variables) & & & & & \\
\hline 
found by \textsc{pdsat} & 0.00036 & 1.02542 & 0.10181 & 0.00523 & 4.64428e+08 \\
(32 variables) & & & & & \\
\hline 
\end{tabular}
\end{table}

Despite the fact that values of the predictive function for the decomposition sets from Fig. \ref{a5-1_sets} are quite close, the value of $s^2$ for the set found by \textsc{pdsat} is significantly less. It means that for this decomposition set the obtained time estimation is more precise according to \eqref{eq4}.

Further we consider the logical cryptanalysis of the Bivium cipher \cite{DBLP:conf/isw/Canniere06}. In Bivium two shifted registers of a special kind are used, first consisting of 93 cells and second consisting of 84 cells. Logical cryptanalysis of Bivium is considered in the following formulation (that was earlier studied in \cite{DBLP:conf/sat/EibachPV08}, \cite{Mcdonald_attackingbivium}, \cite{DBLP:conf/sat/SoosNC09}): based on the known fragment of the keystream one should find 177 bits that correspond to the internal state of the Bivium registers at the start of the keystream generation. In the experiments presented below we considered first 200 bits of the keystream (similar to \cite{DBLP:conf/sat/EibachPV08}).

\begin{figure}[ht]
\centering
\includegraphics[height=2.5cm]{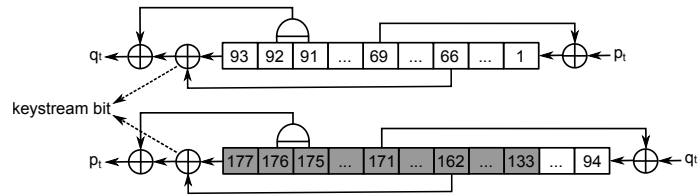}
\caption{Decomposition set of 45 variables for logical cryptanalysis of Bivium from \cite{DBLP:conf/sat/EibachPV08}}
\label{bivium_Ending2_45vars}
\end{figure}

\begin{figure}[ht]
\centering
\includegraphics[width=12cm]{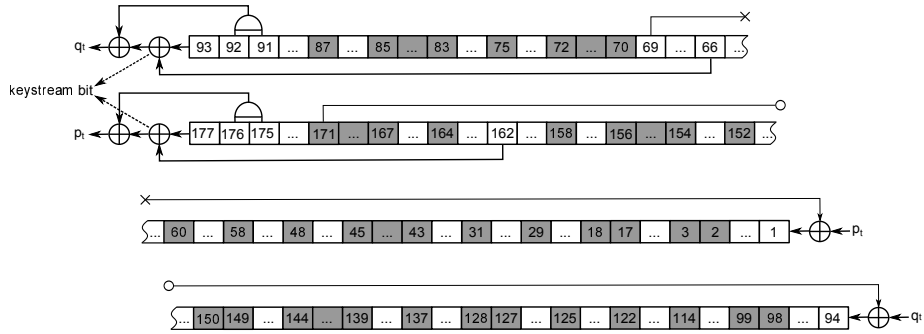}
\caption{Decomposition set of 47 variables for logical cryptanalysis of Bivium that was found by \textsc{pdsat}}
\label{bivium_pdsat_47vars}
\end{figure}

The authors of \cite{DBLP:conf/sat/EibachPV08}, \cite{Mcdonald_attackingbivium}, \cite{DBLP:conf/sat/SoosNC09} presented several variants of decomposition sets. We compared the decomposition set found by \textsc{pdsat} with the results of \cite{DBLP:conf/sat/EibachPV08} since in that work decomposition sets are presented explicitly and those experiments are easy to reproduce. The best decomposition set among the ones presented in \cite{DBLP:conf/sat/EibachPV08} consists of 45 variables and is shown in Fig. \ref{bivium_Ending2_45vars}. In our experiment as an $\tilde{X}_0$ \textsc{pdsat} was given a SUPBS of a CNF encoding the Bivium cryptanalysis problem (177 variables). The decomposition set found by \textsc{pdsat} is shown in Fig. \ref{bivium_pdsat_47vars}.

For the problem of logical cryptanalysis of Bivium we used the \textsc{MiniSat 2.2} solver. In Table \ref{bivium_results} qualitative parameters of the decomposition sets from figures \ref{bivium_Ending2_45vars} and \ref{bivium_pdsat_47vars} are presented.

\begin{table}
\caption{Qualitative parameters of decomposition sets from figures \ref{bivium_Ending2_45vars} and \ref{bivium_pdsat_47vars}.}
\label{bivium_results}
\centering
\begin{tabular}{p{2.2 cm}|p{1.75cm}|p{1.75cm}|p{1.75cm}|p{1.75cm}|p{1.75cm}}

Sets& Min. time & Max. time & Avg. time & $s^2$ & $F\left( \cdot \right)$\\
\hline 
\hline
from \cite{DBLP:conf/sat/EibachPV08}  & 0.05327 & 3.73068 & 0.61090 & 0.09497 & 2.14941e+13 \\
(45 variables) & & & & & \\
\hline 
found by \textsc{pdsat} & 0.00034  &0.83422 & 0.00095 & 0.00031 & 1.33910e+11 \\
(47 variables) & & & & & \\
\hline 
\end{tabular}
\end{table}

In Table \ref{search_info} we present additional information about traversal of the search space during the computational experiments that clearly corroborates the efficiency of the interruption technique described in Subsection 2.3.

\begin{table}
\caption{Information about traversal of the search space by \textsc{pdsat}.}
\label{search_info}
\centering
\begin{tabular}{p{1.7cm}|p{3.3cm}|p{3.3cm}|p{1.2cm}|p{1.2cm}}

&\multicolumn{4}{|c}{Number of points}\\
\cline{2-5}
Experiment & where calculation of $F$ & where calculation of $F$& in $L_1$& in $L_2$\\
& finished& was interrupted&&\\
\hline
\hline
A5/1 & 65 & 11667 & 221 & 11511 \\
\hline
Bivium & 544 & 302991 & 1979 & 301556 \\
\hline
\end{tabular}
\end{table}

It should be noted that in both computational experiments \textsc{pdsat} was provided only with SUPBS of SAT problems. Nevertheless, it managed to find the decomposition sets that are comparable to or better than the ones obtained by manually analyzing the features of the corresponding keystream generators.

\section{Solving hard SAT instances in the volunteer computing project SAT@home}

We use the \textsc{pdsat} program described in Section 3 to plan computational experiments in the volunteer computing project SAT@home. 
SAT@home was launched on the 29th of September 2011 \cite{website:sat@home}. It uses computing resources provided by volunteer PCs to solve hard combinatorial problems that can be effectively reduced to SAT. The project was implemented using the BOINC platform. An experiment that consisted in solving 10 inversion problems of the generator A5/1 was successfully finished in SAT@home on the 7th of May 2012. It should be noted that we considered only instances that cannot be solved using the known rainbow tables \cite{website:rainbow-a5}. 

In that experiment we used the decomposition set from \cite{DBLP:conf/pact/SemenovZBP11}. The computing application was based on a modified version of \textsc{MiniSat-C 1.14.1} (see \cite{DBLP:conf/pact/SemenovZBP11}). First 114 bits of the keystream that correspond to one keystream burst of the GSM protocol were analyzed. On average in order to solve one problem of logical cryptanalysis of A5/1 SAT@home processed about 1 billion SAT problems. 

Since May 2012 SAT@home is occupied in searching for systems of orthogonal Latin squares. During this time we found several pairs of orthogonal diagonal Latin squares of order 10 that are different from the ones published in \cite{Brown93}. 

Characteristics of the SAT@home project as of 8 of February 2013 are (according to BOINCstats\footnote{\url{http://boincstats.com/}}):

\begin{itemize}
	\item 2367 active PCs (active PC in volunteer computing is a PC that sent at least one result in last 30 days) about 80\% of them use Microsoft Windows OSes;
	\item 1299 active users (active user is a user that has at least one active PC);
	\item	versions of the client application: Windows/x86, Linux/x86, Linux/x64;
	\item average real performance: 2,9 teraflops, maximal performance: 6,3 teraflops.
\end{itemize}

The dynamics of the real performance of SAT@home can be seen at the SAT@home performance page\footnote{\url{http://sat.isa.ru/pdsat/performance.php}}.

It should be noted that the estimation for the A5/1 cryptanalysis (see Section 3) obtained with the use of \textsc{pdsat} is close to the average real time spent by SAT@home to solve corresponding SAT problems. With respect to the estimation from Section 3 logical cryptanalysis of Bivium cipher would take about 6 years in SAT@home with its current performance.

\section{Related Work}

Topics related to organization of SAT solving in distributed environments were considered in many papers, for example in \cite{DBLP:journals/pc/BlochingerSK03}, \cite{DBLP:journals/pc/ChrabakhW06}, \cite{DBLP:conf/hvc/HeuleKWB11}, \cite{Hyvarinen11}, \cite{DBLP:conf/sat/HyvarinenJN06}, \cite{DBLP:journals/jsc/ZhangBH96}.

In \cite{Hyvarinen11,DBLP:conf/sat/HyvarinenJN06,DBLP:conf/lpar/HyvarinenJN10,DBLP:conf/cp/HyvarinenJN11} various approaches to partitioning SAT problems were studied. Detailed analysis of a number of problems regarding partitioning approach was presented in \cite{Hyvarinen11}. Also, in \cite{Hyvarinen11} special efficiency functions were introduced to evaluate the quality of a SAT problem partitioning. In our paper, we used predictive functions that are based on different principles.

The authors of \cite{DBLP:conf/lpar/HyvarinenJN10} proposed to use lookahead heuristics \cite{DBLP:series/faia/HeuleM09} to construct SAT problem partitionings. This idea (with significant additions) was implemented in \cite{DBLP:conf/hvc/HeuleKWB11}. During the process of the original problem solving, the distributed solver from \cite{DBLP:conf/hvc/HeuleKWB11} processed hundreds of thousands of SAT instances that correspond to cubes of partitioning that were generated by a lookahead solver on the preprocessing stage.

In \cite{DBLP:journals/grid/SchulzB10} a desktop grid for solving SAT which used conflict clauses exchange via a peer-to-peer protocol was described. Apparently, \cite{DBLP:conf/grid/BlackB11} became the first paper about the use of a desktop grid based on the BOINC platform for solving SAT. Unfortunately, it did not evolve into a full-fledged volunteer computing project. 

The first work that used SAT-solvers for cryptanalysis was \cite{DBLP:journals/jar/MassacciM00}. The authors of \cite{DBLP:conf/sat/EibachPV08}, \cite{Mcdonald_attackingbivium}, \cite{Ecrypt_TOOLS10b}, \cite{DBLP:conf/sat/SoosNC09} presented some estimations of the time required for logical cryptanalysis of the Bivium cipher, obtained using the ideas underlying the Monte-Carlo method. The main novelty of our approach lies in the fact that we consider the process of construction  of the decomposition set with good qualitative parameters as a process of optimisation of predictive function in a special search space.

The most effective method of cryptanalysis of the A5/1 generator is the rainbow method implemented in A5/1 Cracking project. Publicly available rainbow tables \cite{website:rainbow-a5} made it possible to successfully determine the initial state of A5/1 registers based on 8 known bursts of the keystream with probability about 88\% in several seconds on a usual PC. It means that $\sim$12\% of the key space is not covered by these tables. In the SAT@home project we were searching for keys from these 12\% of the key space only. 

Extensive bibliography regarding the use of SAT solvers in searching for combinatorial designs (for example, orthogonal Latin squares) is presented in \cite{DBLP:series/faia/Zhang09}.

\section{Conclusions}
In this paper, we proposed a method for estimating time to solve SAT in distributed computing environments. It uses the Monte Carlo method to statistically estimate the quality of partitioning of the original SAT problem. To search for a partitioning with good quality a special tabu search algorithm was used. The proposed method was used to obtain an approximate time required to do logical cryptanalysis of the well-known ciphers A5/1 and Bivium. Ten problems of logical cryptanalysis of the A5/1 generator, that could not be solved using the known rainbow tables \cite{website:rainbow-a5}, were successfully solved in the volunteer computing project SAT@home \cite{website:sat@home} that was developed and is maintained by the authors.

\subsubsection*{Acknowledgments.} 
Authors thank Stepan Kochemazov for numerous valuable comments that allowed us to significantly improve the quality of the paper, Alexey Ignatiev for constructive feedback and helpful discussions, Mikhail Posypkin and Nikolay Khrapov for their help in developing and administering of the SAT@home project, Karsten Nohl for detailing the rainbow method used in the A5/1 Cracking Project. Also we express our gratitude to all the users participating in the SAT@home project for their dedication and enthusiasm. This work was supported by Russian Foundation for Basic Research, grant 11-07-00377a. Oleg Zaikin also acknowledges support from President of Russian Federation grant for young scientists SP-1855.2012.5. 

\bibliographystyle{splncs03}
\bibliography{refs}

\end{document}